\pdfoutput=1

\documentclass[11pt]{article}
\usepackage{amssymb}

\usepackage[final]{acl}

\usepackage{times}
\usepackage{latexsym}
\usepackage{amsmath}
\usepackage{multirow}
\usepackage[normalem]{ulem}
\useunder{\uline}{\ul}{}
\usepackage{graphicx}
\usepackage{subcaption}

\usepackage[T1]{fontenc}

\usepackage[utf8]{inputenc}

\usepackage{microtype}

\usepackage{inconsolata}

\usepackage{graphicx}

\title{MaskCD: Mitigating LVLM Hallucinations by Image Head Masked Contrastive Decoding}

\author{
  Jingyuan Deng$^{1}$ \quad Yujiu Yang$^{1}$\thanks{Corresponding author} \\
  $^{1}$ Tsinghua Shenzhen International Graduate School, Tsinghua University \\
  \texttt{deng-jy24@mails.tsinghua.edu.cn}
}

\begin{document}
\maketitle
\begin{abstract}
Large vision-language models (LVLMs) have shown remarkable performance in visual-language understanding for downstream multimodal tasks. While their capabilities are improving, problems emerge simultaneously. Among those problems, the hallucinations have attracted much attention, which stands for the phenomenon where LVLMs generate contradictory content to their input visual and text contents. Many approaches have been proposed to deal with this issue, such as contrastive decoding and attention manipulation. However, contrastive decoding methods struggle in constructing appropriate contrastive samples, and attention manipulation methods are highly sensitive, lacking stability. In this work, we propose image head \textbf{Mask}ed \textbf{C}ontrastive \textbf{D}ecoding (\textbf{MaskCD}). Our approach utilizes the "image heads" in LVLMs, masking them to construct contrastive samples for contrastive decoding. We evaluated MaskCD on LLaVA-1.5-7b and Qwen-VL-7b, using various benchmarks such as CHAIR, POPE, AMBER and MME. The results demonstrate that MaskCD effectively alleviates the phenomenon of hallucinations and retains the general capabilities of LVLMs. Corresponding resources could be found at: \href{https://github.com/Deng-Jingyuan/MaskCD}{https://github.com/Deng-Jingyuan/MaskCD} 

\end{abstract}

\section{Introduction}



\begin{figure}[ht]
    \centering
    \includegraphics[width=\linewidth]{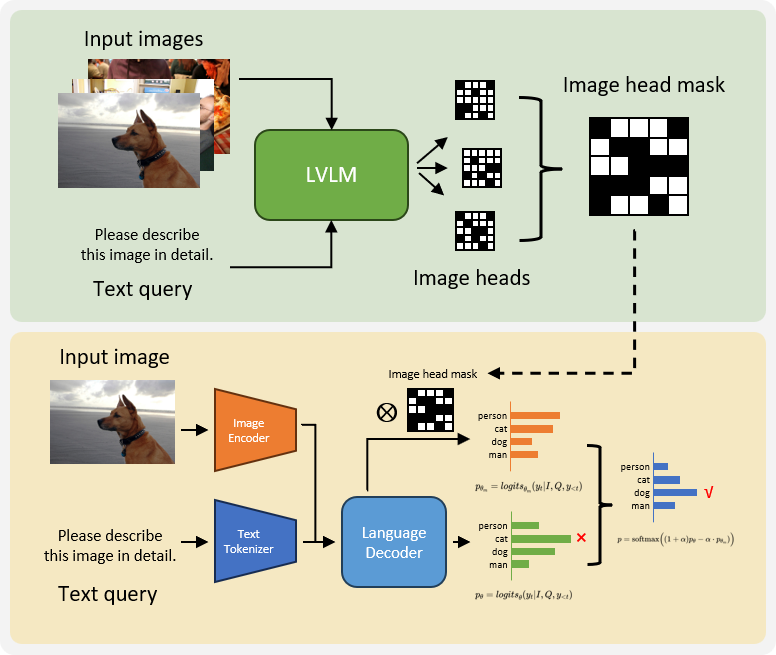}
    \caption{\textbf{Pipeline of MaskCD.}The upper part shows the first step. The image head mask is constructed by querying LVLM with images and prompt texts. Then, the lower part shows how to use the image head mask in the process of contrastive decoding.}
    \label{fig: mcd}
\end{figure}

Large Language Models (LLMs) (\citealp{DBLP:conf/nips/BrownMRSKDNSSAA20},\citealp{DBLP:journals/corr/abs-2303-08774}) have achieved remarkable success in understanding human instructions and performing diverse tasks. Building on this progress, recent efforts have extended LLMs to develop Large Vision-Language Models (LVLMs) (\citealp{DBLP:journals/corr/abs-2308-12966},\citealp{DBLP:conf/icml/0008LSH23},\citealp{DBLP:conf/nips/Dai0LTZW0FH23},\citealp{DBLP:journals/corr/abs-2304-10592},\citealp{DBLP:journals/corr/abs-2311-04257},\citealp{DBLP:journals/corr/abs-2310-03744}), which integrate visual and textual modalities for multimodal reasoning. Although researchers have already achieved remarkable success in applying LVLMs into several tasks, problems have emerged as well. Within these problems, the hallucination (\citealp{DBLP:journals/corr/abs-2403-05262},\citealp{DBLP:conf/emnlp/KamathHC23a},\citealp{DBLP:conf/emnlp/LiDZWZW23}) has attracted significant attention.

The hallucination of LVLMs is a phenomenon in which models tend to generate contradictory contents for the inputs, especially images. This may manifest as generating non-existent objects, mistakenly described attributes, or non-sense sentences. All kinds of hallucinations enormously lower users' trust in the model and even cause fatal damage when applied in real-world tasks like auto-driving and medical image processing.

\begin{figure*}[ht]
  \hfill
  \includegraphics[width=0.45\linewidth]{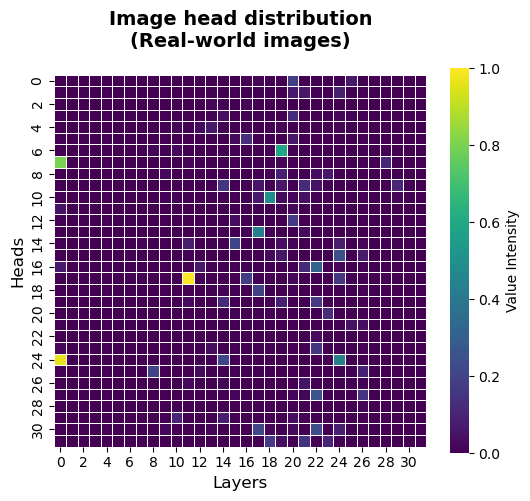} \hfill
  \includegraphics[width=0.45\linewidth]{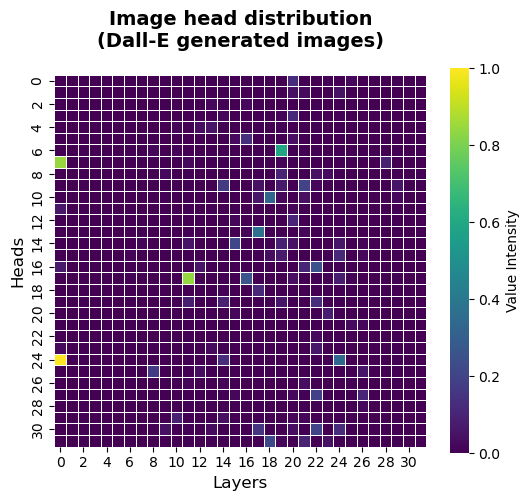} \hfill
  \hfill
  \caption {\textbf{Visualization of image heads in LLaVA-1.5-7b.} The left figure shows the image head distribution of real-world images, while the right one represents the results of Dall-E generated artificial images. It is evident that there are certain heads that tend to pay high attention on image tokens, therefore we name them with "image head".}
  \label{fig: image heads}
\end{figure*}

To mitigate the hallucination phenomenon, researchers have promoted multiple methods, which could be classified into two main categories according to whether training is needed. Firstly, training-involved methods (\citealp{DBLP:conf/naacl/LeePJS24},\citealp{DBLP:conf/cvpr/ChenSCJD24},\citealp{DBLP:conf/iclr/LiuLLWYW24}) collect massive delicately-constructed data to fine-tune or post-train the LVLMs, so as to teach models to generate less hallucinated content. It features with a high-cost of computation resources and massive human labor. On the contrary, training-free methods are developed to alleviate hallucination at a lower cost. Prevailing methods include contrastive decoding (CD) (\citealp{DBLP:conf/cvpr/LengZCLLMB24},\citealp{DBLP:conf/cvpr/FaveroZTCPASS24},\citealp{DBLP:journals/corr/abs-2405-17821}) and attention manipulation (\citealp{DBLP:journals/corr/abs-2503-08342},\citealp{DBLP:conf/cvpr/HuangDZ0H0L0Y24},\citealp{DBLP:conf/eccv/LiuZC24}). CD methods need an injured input as the bad sample, whose output logits will be subtract from the original one. It would take twice times of inference cost, once for the original input, the other for the injured one. But with delicately constructed bad samples, CD methods have presented prominent performance in mitigating the LVLMs' hallucination phenomenon. Recently, with the progress in understanding models' inner working mechanisms, methods of attention manipulation have come up. Specifically, abnormal attention map phenomena, such as excessive local attention in the attention sink phenomenon (\citealp{DBLP:conf/cvpr/FaveroZTCPASS24},\citealp{DBLP:conf/iclr/XiaoTCHL24}), will be directly reduced or redistributed. This method features in better align visual and text modal, enabling models to better and truly utilize visual information.

Although the need for training has been eliminated, both the CD and attention manipulation methods have their drawbacks. The performance of CD methods is heavily depended on the quality of the constructed bad sample. If the injured sample still contains a lot of useful information, then the contrast operation may even cause worse results. For the attention manipulation methods, models are highly sensitive to changes in the attention score and are not as stable as CD methods in terms of overall testing scenarios.

Therefore, we hope to construct high-quality bad samples through the angle of attention distribution, thereby combining the advantages of CD methods and attention manipulation methods, balancing stability and hallucination-mitigating performance. To observe the model's attention preference to images at head-level, we randomly select 500 images in the validation set of COCO 2014(\citealp{DBLP:conf/eccv/LinMBHPRDZ14}) and 500 Dall-E generated artificial images from MMrel (\citealp{DBLP:journals/corr/abs-2406-09121}). Put them into LLaVA-1.5-7b (\citealp{DBLP:journals/corr/abs-2310-03744}) with the prompt text "Please describe this image in detail." and record the sum of the attention scores obtained by each head in each layer of the model for each token generation. Finally, under different thresholds, the number of times each head pays excessive attention to the image token during the generation of each token is calculated.

The normalized result is visualized in Figure \ref{fig: image heads}. We observed that whether real-world images or AI-generated images, there are certain heads in LLaVA-1.5-7b that prefer to give image tokens comparably high attention scores. Since they present an inclined focus on visual information, we name them "image heads". Given that the essence of CD methods is making the subtracted samples contain only invalid information as much as possible, we choose to mask these image heads to construct bad samples, so as to prevent the bad samples from accessing useful visual information more precisely. 

In this way, we proposed image head Masked Contrast Decoding (MaskCD), which uses image head attention masks to construct delicate bad samples and attain significant hallucination-mitigating performance.

Our contributions can be summarized as follows:
\begin{itemize}
    \item We identify "image heads" in LVLMs that disproportionately attend to image tokens.
    \item We propose MaskCD, a contrastive decoding method that features in using image head masking to construct degraded visual inputs.
    \item We demonstrate, through extensive experiments, that MaskCD outperforms existing hallucination mitigation methods across multiple benchmarks while preserving general model capabilities.
\end{itemize}

\section{Related Work}

\subsection{Large Vision-Language Model}

Recently, efforts have been made to enhance Large Vision-Language Models, aiming to equip LLMs with the ability to process visual information like images or videos.
LVLMs are typically constructed by three components: a visual encoder to extract visual features, a modality connection module to bridge visual and text modal, and an LLM for further tasks. The visual encoder and LLM are typically fixed pretrained models; common choices are CLIP model (\citealp{DBLP:conf/icml/RadfordKHRGASAM21}) variants for the visual encoder, and LLaMA (\citealp{DBLP:journals/corr/abs-2302-13971}) or Vicuna (\citealp{vicuna_blog}) for the LLM.

Research focuses on optimizing modality connection modules, so as to better utilize visual and text information at the same time. Different connection modules lead to different LVLM types: cross-attention module in Flamingo(\citealp{DBLP:conf/nips/AlayracDLMBHLMM22}), Q-former in BLIP-2(\citealp{DBLP:conf/icml/0008LSH23}), and simple linear layer in LLaVA(\citealp{DBLP:journals/corr/abs-2310-03744}) model series.

\subsection{Hallucination in LVLMs}

Multimodal hallucination phenomenon, typically presented as LVLM generates inconsistent content from the input, especially those it contradictory with visual information. For example, in the image captioning task, LVLM may generate objects that do not exist in the input images (\citealp{DBLP:conf/emnlp/LiDZWZW23}), or mistakenly describe attribution of existing objects like counts, color and spatial relationship (\citealp{DBLP:conf/emnlp/KamathHC23a}).

The methods for alleviating LVLM hallucinations can be classified according to whether training is required. Training-involved methods typically uses constructed data to fine-tune or post-train LVLMs. For example, \citet{DBLP:journals/corr/abs-2309-02301}, \citet{DBLP:journals/corr/abs-2306-14565} uses contrastive question-answer pairs to fine-tune LVLMs, and \citet{DBLP:conf/acl/SunSCLLSGGWYKD24} employs Reinforcement Learning from Human Feedback (RLHF) to enhance multimodal connections. Training-free methods are prevailed by contrastive decoding and attention manipulation. The core of CD methods is constructing bad samples that contain useful information as less as possible. Different constructing means like image editing(\citealp{DBLP:journals/corr/abs-2405-17821},\citealp{DBLP:conf/cvpr/LengZCLLMB24}), text editing(\citealp{DBLP:conf/acl/WangPDB24}) and model bias(\citealp{DBLP:journals/corr/abs-2402-18476}) are developed to achieve this goal. Attention manipulation method would reduce(\citealp{DBLP:conf/cvpr/HuangDZ0H0L0Y24}) or redistribute(\citealp{DBLP:journals/corr/abs-2503-08342}) excessive attention scores, so as to steer LVLMs to pay more attention to visual information.

CD methods perform well in hallucination-mitigating tasks but are highly dependent on the quality of the bad samples constructed. If the injured sample still contains a lot of useful information, the contracting operation may cause an even worse result. Attention manipulation methods cost fewer computation resources but are highly sensitive to parameters, presenting unstable performances. Our research constructs bad samples from the perspective of attention, filtering out useful information so that the bad samples only carry the information that needs to be offset, thereby achieving a stable and high-quality effect.

\section{Methodology}

\subsection{Task Formation}
Typically, LVLMs aim to generate proper text outputs from multimodal inputs, especially combined visual and textual data. The visual encoder extracts visual features, then passes them to the modal connection module, where visual features are mapped into the text semantic space. The mapped features are combined with textual tokens, either through concatenation(\citealp{DBLP:journals/corr/abs-2310-03744}) or cross-modal fusion(\citealp{DBLP:conf/nips/Dai0LTZW0FH23}). The final combined features are then passed into the LLM to generate outputs autoregressively. Formally, given an input image $I$, corresponding question text $Q$, and already generated tokens $y_{<t}$, the next token $y_t$ is decoded according to the probability distribution:
\begin{equation}
  \label{eq:eq1}
  p(y_t)=p_{\theta}(y_t\ |\ I,Q,y_{<t})
\end{equation}
where $\theta$ represents the parameters of the LVLM. The goal of hallucination mitigation is to make output sequences contain less contradictory content.

\subsection{Formulating image heads masks} \label{subsec:mask formulation}
LLMs in prevailing LVLMs are most decoder-only, use an attention mechanism to capture the blending of textual and visual features. Formally, the attention matrix $A$ is calculated by:
\begin{equation}
  \label{eq:eq2}
  A=\text{softmax}(\frac{Q\cdot K^T}{\sqrt{d_k}})
\end{equation}
where $Q$ and $K$ denotes queries and keys respectively, $d_k$ represents the dimension of key vectors. Each row of the attention matrix $A$ indicates the proportion of attention that the current token has invested in the previously generated token. We believe that the higher the sum of the values obtained by the image tokens in the attention matrix is, the more attention the visual information will receive.

There are multiple attention heads in each layer of the LLM, each of which calculates its own attention matrix. We randomly selected 500 images from the validation set of COCO 2014(\citealp{DBLP:conf/eccv/LinMBHPRDZ14}), input them into LVLMs with the text 'Please describe this image in detail', then record the sum of the attention scores for the image token in the attention matrix of each head in each layer of the model when each token is generated. With a threshold $\tau$, we obtain the attention head matrix where each element represents how many times this attention head has paid over-threshold attention proportion to image tokens (as shown in Figure \ref{fig: image heads}). After normalization, the non-zero attention heads in the attention head matrix are named image heads. Lastly, by masking the selected image heads, the image head mask is constructed. For more information, please refer to appendix \ref{sec:appendix}, including the formalized description of MaskCD and more details about image head selection.

\begin{table}[t]
\centering
\resizebox{\columnwidth}{!}{%
\begin{tabular}{cccc}
\hline
Model                         & $\tau$ & \# image heads & proportion \\ \hline
\multirow{6}{*}{LLaVA-1.5-7b} & 0.95                & 192                   & 18.75\%    \\
                              & 0.9                 & 238                   & 23.24\%    \\
                              & 0.8                 & 315                   & 30.76\%    \\
                              & 0.7                 & 364                   & 35.55\%    \\
                              & 0.6                 & 424                   & 41.41\%    \\
                              & 0.5                 & 506                   & 49.41\%    \\ \hline
\multirow{4}{*}{Qwen-VL-7b}   & 0.99                & 248                   & 24.22\%    \\
                              & 0.975               & 317                   & 30.96\%    \\
                              & 0.95                & 395                   & 38.57\%    \\
                              & 0.9                 & 473                   & 46.19\%    \\ \hline
\end{tabular}%
}
\caption{\textbf{The number and proportion of image heads corresponding to the variation of $\tau$.} $\tau$ represents the threshold of considering "high" attention scores paid on image tokens of a head.}
\label{tab:head num}
\end{table}

Apparently, the number of image heads varies with the change of $\tau$. Table \ref{tab:head num} shows the number of image heads of LLaVA-1.5-7b and Qwen-VL-7b(\citealp{DBLP:journals/corr/abs-2308-12966}) given different threshold $\tau$ values. Intuitively, if the threshold is too high and too few bad samples are masked, then useful information will still be contained in the bad samples; If the threshold is too low, causing the heads that do not pay much attention to the image to be masked as well, the reduction effect of the CD method on semantic information will be weakened. Therefore, choosing the appropriate threshold is an important issue. 

\subsection{MaskCD} \label{subsec: maskcd}
When using an image head mask to construct a bad sample, the masked heads' attention output will be set to zero. Since this method is equivalent to setting the parameters of the corresponding head to zero, we use $\theta_m$ to represent the model where the image heads are masked. However, in actual operation, only the attention value is changed; no model parameter will be modified.

Then MaskCD is formulated as equation \ref{eq:MaskCD}:
\begin{equation}
  \label{eq:MaskCD}
  \begin{split}
    p(y_t) &= \text{softmax}\Bigl((1+\alpha) \cdot \text{logits}_\theta(y_t|I,Q,y_{<t}) \\
           &\quad - \alpha \cdot \text{logits}_{\theta_m}(y_t|I,Q,y_{<t})\Bigr)
  \end{split}
\end{equation}
where $logits$ represents the value of $p(y_t|I,Q,y_{<t})$ before softmax operation. $\alpha$ is a hyperparameter that controls the intensity of contrast. 

By subtracting the output logits of bad samples from the original ones, MaskCD enables the final output logits to utilize only the truly useful visual and textual information as much as possible, thereby alleviating the hallucination phenomenon of LVLMs.

\section{Experiment Settings}
\subsection{Benchmarks}
\paragraph{CHAIR} The Caption Hallucination Assessment with Image Relevance (CHAIR) (\citealp{DBLP:conf/emnlp/RohrbachHBDS18}) is a widely used metric for evaluating object hallucination in image captioning tasks. CHAIR is used to measure the hallucination proportion of the model's generated texts. It evaluates hallucination on two aspects: CHAIR$_S$ and CHAIR$_I$. The former calculates the proportion of sentences containing hallucinations at the sentence level, while the latter computes the hallucinated ratio at the object level. The two metrics can be formulated as:
\begin{equation}
  \label{eq:CHAIR}
  \begin{split}
    \text{CHAIR}_S &= \frac{|\{\text{sentences w/ hallucinated objects}\}|}{|\{\text{all captions sentences}\}|} \\
    \text{CHAIR}_I &= \frac{|\{\text{hallucinated objects}\}|}{|\{\text{all mentioned objects}\}|} \\
  \end{split}
\end{equation}

We randomly selected 500 images from the validation set of COCO 2014(\citealp{DBLP:conf/eccv/LinMBHPRDZ14}) and used the prompt "Please describe this image in detail." to obtain the generated captions.

\paragraph{POPE} The Polling-based Object Probing Evaluation (POPE) (\citealp{DBLP:conf/emnlp/LiDZWZW23}) is a benchmark for assessing object hallucination. LVLMs are required to answer formatted questions like "Is there a <object> in the image?" with "Yes" or "No". The answers' yes-no ratio is designed to be 50\% for each response. The complete POPE test is divided into three splits: random, popular and adversarial, in which missing objects are randomly selected, most frequently occurring in the dataset, and highly correlated with those present in the image, respectively.

We choose MSCOCO dateset for POPE evaluation. The key evaluation metrics are: Accuracy, Precision, Recall, and F1 score.

\paragraph{AMBER} AMBER(\citealp{wang2023amber}) is An LLM-free Multi-dimensional Benchmark for MLLMs hallucination evaluation, which can be used to evaluate both generative task and discriminative task including existence, attribute and relation hallucination.

\paragraph{MME} The Multimodal Large Language Model Evaluation (MME) (\citealp{DBLP:journals/corr/abs-2306-13394}) assesses LVLMs using set of comprehensive metrics. MME benchmark contains 14 subsets, so as to evaluate LVLMs' general capabilities. Following the methodologies of (\citealp{DBLP:journals/corr/abs-2310-16045}), when presenting the test results of all subsets of MME, we divide them into two groups: hallucination and non-hallucination. The hallucination group includes "existence", "counts", "color" and "position", which evaluate LVLMs at the object and attribute level,s respectively.

\subsection{Models}
\textbf{LVLM Models}    We select LLaVA-1.5-7b and Qwen-VL-7b for evaluation. Each model utilizes Vision Transformer (ViT) as the backbone of its visual encoder, but employs different modal connection modules and LLMs. LLaVA-1.5-7b directly projects visual embeddings into semantic space through multi-layer perception(MLP), while Qwen-VL-7b utilizes a position-aware vision-language adapter to compress image features. As for the LLM part, LLaVA-1.5-7b utilizes vicuna as LLM backbone, while Qwen-VL-7b's counterpart is Qwen(\citealp{DBLP:journals/corr/abs-2309-16609}). The LLM backbones are both constructed by 32 layers of decoder blocks, and each layer contains 32 heads, resulting in 1024 heads in total.

\subsection{Baseline Methods}
We compare MaskCD with three classic and effective hallucination mitigating methods: \textbf{VCD}(\citealp{DBLP:conf/cvpr/LengZCLLMB24}) uses random Gaussian noise to contaminate the original image, reducing the valid information it contains and thus serving as a bad sample. \textbf{M3ID}(\citealp{DBLP:conf/cvpr/FaveroZTCPASS24})   deletes the image for the bad sample input, and slightly changes the contrastive decode function. The above methods all belong to the CD category. \textbf{OPERA}(\citealp{DBLP:conf/cvpr/HuangDZ0H0L0Y24}) takes advantage of the attention sink phenomenon, punishes overly concentrated attention, and combines it with a retrospection-allocation strategy. It is an attention manipulation method based on a beam search strategy. MaskCD is a CD method that only masks the model's inner values to construct bad samples, distinguishing it from other CD-class methods.

\section{Result and Analysis}
\subsection{Overall Result}

\begin{table*}[!ht]
\centering
\resizebox{\textwidth}{!}{
\begin{tabular}{lcccccccc}
\hline
Method               & \multicolumn{4}{c}{LLaVA-1.5-7b}                                       & \multicolumn{4}{c}{Qwen-VL-7b}                       \\
\multicolumn{1}{c}{} & CHAIR\_s ↓ & CHAIR\_i ↓ & Precision  & F1                              & CHAIR\_s ↓ & CHAIR\_i ↓ & Precision  & F1            \\ \hline
Baseline             & 50.20      & 15.40      & 72.10      & \multicolumn{1}{c|}{73.50}      & 50.8       & 17.4       & 68.3       & 63.0          \\
VCD                  & 55.6       & 16.4       & 71.7       & \multicolumn{1}{c|}{75.2}       & 48.4       & 16.7       & 68.1       & 64.6          \\
M3ID                 & 55.4       & 15.5       & 70.6       & \multicolumn{1}{c|}{75.7}       & {\ul 39.8} & {\ul 8.8}  & {\ul 77.2} & {\ul 75.6}    \\
OPERA                & 45.8       & {\ul 13.5} & {\ul 76.6} & \multicolumn{1}{c|}{{\ul 77.8}} & 42.3       & 11.8       & 75.5       & \textbf{76.3} \\
MaskCD & \textbf{40.6} & \textbf{10.8} & \textbf{79.1} & \multicolumn{1}{c|}{\textbf{78.2}} & \textbf{12.4} & \textbf{8.6} & \textbf{88.5} & 64.3 \\ \hline
\end{tabular}%
}
\caption{\textbf{Results on benchmark CHAIR.} CHIAR\_s and CHAIR\_i are hallucination ratio evaluation metrics, lower scores represent better performances. The Baseline method denotes the standard decoding. The best performances within each setting are bolded. Comparable but not the best performances are underlined.  }
\label{tab:CHAIR main}
\end{table*}

\begin{table*}[!ht]
\centering
\begin{tabular}{cccccccc}
\hline
\multirow{2}{*}{\textbf{Setup}}       & \multirow{2}{*}{\textbf{Method}} & \multicolumn{3}{c}{\textbf{LLaVA-1.5-7b}} & \multicolumn{3}{c}{\textbf{Qwen-vl-7b}} \\ \cline{3-8} 
                                  &                & Acc. ↑         & Recall ↑       & F1 ↑           & Acc. ↑         & Recall ↑       & F1 ↑           \\ \hline
\multirow{6}{*}{\textit{random}}  & Baseline       & 82.90          & 72.07          & 80.82          & {\ul 81.97}    & 77.67          & {\ul 81.16}    \\
                                  & VCD            & 85.57          & 76.27          & 84.09          & 76.20          & \textbf{81.73} & 77.45          \\
                                  & M3ID           & 85.27          & 74.67          & 85.52          & 74.60          & 69.67          & 73.28          \\
                                  & OPERA          & \textbf{89.30} & \textbf{89.00} & \textbf{89.27} & 66.33          & \textbf{81.73} & 77.45          \\
                                  & MaskCD         & {\ul 88.77}    & {\ul 87.47}    & {\ul 88.62}    & \textbf{87.77} & {\ul 79.47}    & \textbf{86.66} \\ \hline
\multirow{6}{*}{\textit{popular}} & Baseline       & 81.10          & 74.27          & 79.22          & {\ul 80.20}    & 78.40          & {\ul 79.84}    \\
                                  & VCD            & 83.67          & 72.34          & 82.36          & 72.30          & \textbf{81.80} & 74.70          \\
                                  & M3ID           & 83.60          & 73.77          & 81.99          & 72.07          & 70.33          & 71.57          \\
                                  & OPERA          & \textbf{85.93} & \textbf{87.96} & \textbf{86.86} & 66.77          & 73.67          & 68.91          \\
                                  & MaskCD         & {\ul 85.67}    & {\ul 87.53}    & {\ul 85.83}    & \textbf{86.57} & {\ul 79.40}    & \textbf{85.53} \\ \hline
\multirow{6}{*}{\textit{adversarial}} & Baseline                         & 78.60        & 72.35        & 77.10       & {\ul 78.43}   & 78.60   & {\ul 78.47}   \\
                                  & VCD            & {\ul 81.07}    & 76.24          & 80.11          & 71.57          & \textbf{83.07} & 71.50          \\
                                  & M3ID           & \textbf{81.57} & 73.36          & 80.20          & 71.83          & 70.67          & 71.50          \\
                                  & OPERA          & 79.00          & \textbf{88.03} & {\ul 80.91}    & 67.50          & 73.67          & 69.38          \\
                                  & MaskCD         & 79.63          & {\ul 87.53}    & \textbf{81.12} & \textbf{83.40} & {\ul 79.27}    & \textbf{82.68} \\ \hline
\multirow{6}{*}{\textit{All}}     & Baseline       & 80.87          & 72.90          & 79.05          & {\ul 80.20}    & 78.22          & {\ul 79.82}    \\
                                  & VCD            & 83.44          & 74.95          & 82.19          & 73.36          & \textbf{82.20} & 75.55          \\
                                  & M3ID           & 83.48          & 73.93          & 82.57          & 72.83          & 70.22          & 72.12          \\
                                  & OPERA          & {\ul 84.74}    & \textbf{88.33} & \textbf{85.68} & 66.87          & 73.67          & 68.97          \\
                                  & MaskCD         & \textbf{84.88} & {\ul 88.20}    & {\ul 85.48}    & \textbf{85.91} & {\ul 79.38}    & \textbf{84.96} \\ \hline
\end{tabular}
\caption{\textbf{Results on benchmark POPE.} The Baseline method denotes the standard decoding. The best performances within each setting are bolded. Comparable but not the best performances are underlined.}
\label{tab:POPE main}
\end{table*}

\begin{table}[h!]
\centering
\resizebox{\columnwidth}{!}{%
\begin{tabular}{cccccl}
\hline
Method   & \multicolumn{2}{c}{Object\-level}  & \multicolumn{2}{c}{Attribute\-level} & Total           \\ \cline{2-5}
         & existence       & count           & position         & color            &                 \\ \hline
Baseline & {\ul 180.00}    & 101.67          & 100.00           & 153.33           & 535.00          \\
VCD      & 175.00          & 106.67          & 111.67           & 146.67           & 540.01          \\
M3ID     & {\ul 180.00}    & 101.67          & 105.00           & \textbf{158.33}  & 545.00          \\
OPERA    & \textbf{195.00} & {\ul 148.33}    & {\ul 128.33}     & {\ul 155.00}     & {\ul 626.66}    \\
MaskCD   & \textbf{195.00} & \textbf{168.33} & \textbf{133.33}  & 150.00           & \textbf{646.66} \\ \hline
\end{tabular}%
}
\caption{\textbf{Results on benchmark MME (four hallucination subsets) of LLaVA-1.5-7b.} The Baseline method denotes the standard decoding. The best performances within each setting are bolded. Comparable but not the best performances are underlined.}
\label{tab:MME hal llava}
\end{table}

\begin{table}[h!]
\centering
\resizebox{\columnwidth}{!}{%
\begin{tabular}{cccccl}
\hline
Method   & \multicolumn{2}{c}{Object-level} & \multicolumn{2}{c}{Attribute-level} & Total        \\ \cline{2-5}
         & existence       & count          & position         & color            &              \\ \hline
Baseline & 105.00          & 83.33          & 50.00            & {\ul 136.67}     & 375.00       \\
VCD      & 86.67           & 91.67          & 41.67            & 111.67           & 346.00       \\
M3ID     & \textbf{118.33} & \textbf{98.33} & {\ul 51.67}      & 125.00           & 331.68       \\
OPERA    & 93.67           & 84.33          & 46.67            & 121.33           & {\ul 393.33} \\
MaskCD   & {\ul 111.67}    & {\ul 95.00}    & \textbf{65.00}   & \textbf{138.33}  & \textbf{410} \\ \hline
\end{tabular}%
}
\caption{\textbf{Results on benchmark MME (four hallucination subsets) of Qwen-VL-7b.} The Baseline method denotes the standard decoding. The best performances within each setting are bolded. Comparable but not the best performances are underlined.}
\label{tab:MME hal qwen}
\end{table}

\begin{figure*}[!ht]
  \hfill
  \includegraphics[width=0.36\linewidth]{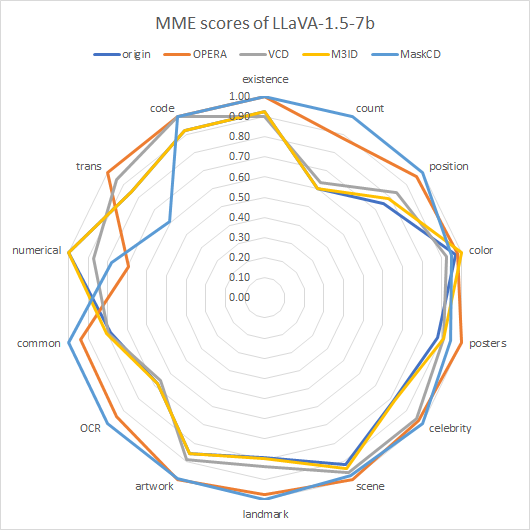} \hfill
  \includegraphics[width=0.36\linewidth]{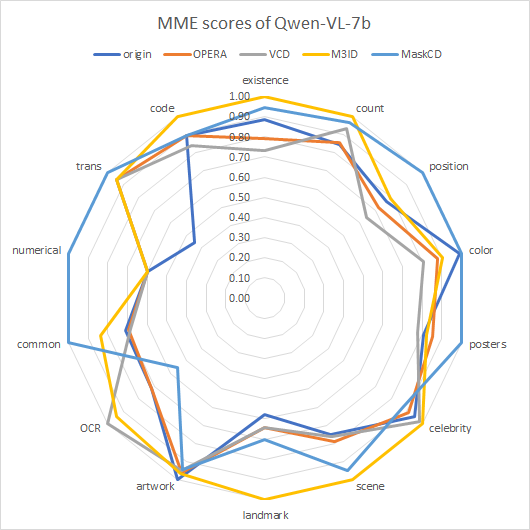} \hfill
  \hfill
  \caption {\textbf{Visualization of MME scores of LLava-1.5-7b(left) and Qwen-VL-7b(right).} Scores are normalized by dividing maximum score of each subset.}
  \label{mme all}
\end{figure*}

\begin{table}[]
\centering
\resizebox{\columnwidth}{!}{%
\begin{tabular}{llcccclcccc}
\hline
Method               &  & \multicolumn{9}{c}{LLaVA-1.5-7b}                                                                                               \\
\multicolumn{1}{c}{} &  & \multicolumn{4}{c}{Generative}                              &  & \multicolumn{4}{c}{Discriminative}                            \\
\multicolumn{1}{c}{} &  & CHAIR ↓      & Cover         & Hal ↓         & Cog ↓        &  & Accuracy      & Preicision    & Recall        & F1            \\ \hline
Baseline             &  & 9.2          & 41.3          & 29.2          & 3.7          &  & 65.7          & 83.2          & 64.7          & 73.29         \\
VCD                  &  & {\ul 8.1}    & 44.2          & {\ul 28.6}    & 3.1          &  & 68.3          & \textbf{85.8} & 65.2          & 74.09         \\
M3ID                 &  & \textbf{7.9} & {\ul 45.3}    & \textbf{28.3} & \textbf{2.8} &  & 69.7          & {\ul 84.9}    & 64.8          & 73.50         \\
OPERA                &  & 8.3          & 43.1          & 31.2          & {\ul 2.9}    &  & {\ul 76.0}    & 79.2          & {\ul 83.8}    & {\ul 81.44}   \\
MaskCD               &  & 8.7          & \textbf{48.6} & 34.5          & 3.2          &  & \textbf{77.8} & 81.4          & \textbf{86.8} & \textbf{84.0} \\ \hline
\end{tabular}%
}
\caption{\textbf{Results on benchmark AMBER on LLaVA-1.5-7b.} The Baseline method denotes the standard decoding. The best performances within each setting are bolded. Comparable but not the best performances are underlined.}
\label{tab:AMBER main}
\end{table}

\paragraph{CHAIR}  Table \ref{tab:CHAIR main} shows the overall results for CHAIR evaluation. MaskCD gained evidently better performance compared with other methods. Specifically, MaskCD lowers CHAIR\_s and CHAIR\_i by 19.12\% and 29.87\% for LLaVA-1.5-7b, and achieves 75.59\% and 50.57\% decrease for Qwen-VL-7b. This indicates that our proposed image heads masks are quite effective in hallucination mitigating. Moreover, MaskCD outperforms VCD and M3ID, demonstrating that the bad samples constructed by masking the image head contain less effective information, thereby achieving better results among similar CD methods.

\paragraph{POPE} Table \ref{tab:POPE main} shows the evaluation results of POPE benchmark. For both LLaVA-1.5-7b and Qwen-VL-7b, MaskCD represents comparable performance with OPERA, outperforms the baseline and other CD methods, indicating its excellence in hallucination alleviation. Furthermore, as taking computational cost into account, MaskCD achieves a similar performance effect with a computational cost lower than that of OPERA, demonstrating its excellence.

\paragraph{AMBER} Table \ref{tab:AMBER main} shows the evaluation results of benchmark AMBER on LLaVA-1.5-7b. MaskCD achieves comparably good performance, especially in discriminative tasks. 

\paragraph{MME} Table \ref{tab:MME hal llava} and Table \ref{tab:MME hal qwen} shows the results on four hallucination-related MME subsets for LLaVA-1.5-7b and Qwen-VL-7b, respectively. MaskCD achieves best performances on every single subset for LLaVA-1.5-7b and on attribute-level subsets for Qwen-VL-7b. The evaluation on object-level subsets of Qwen-VL-7b also achieves the second-best results, representing the effectiveness of MaskCD in alleviating hallucinations. Meanwhile, Figure \ref{mme all} shows the overall results for all 14 subsets of the MME benchmark. It is evident that besides the capability of mitigating hallucination, MaskCD also retains or even partially improves the model's ability in general evaluation.

\subsection{Ablation Study}

In this subsection, we present ablation studies to examine the impact of mask selecting and other hyper-parameters. We conduct these experiments with LLaVA-1.5-7b.

\paragraph{Mask selection} To demonstrate the necessity of masking the image heads rather than other heads, for each image head mask in the settings, we randomly select an equal number from other heads to form a random mask. The method of using these random masks for MaskCD is denoted as MaskCD\_r. Table \ref{tab:ablation-chair}, \ref{tab:ablation pope}, and \ref{tab:ablation-mme} show the performance of MaskCD and MaskCD\_r on CHAIR, POPE and MME, respectively. The results show that masking random heads also has a slight effect on alleviating hallucinations, but it cannot compete with the results of masking image heads. It indicates that the image heads indeed contain more useful and necessary information, so it is rational to mask them rather than other heads.

\begin{table}[!t]
\centering
\resizebox{\columnwidth}{!}{%
\begin{tabular}{lcccc}
\hline
\multicolumn{1}{c}{Method} & CHAIR\_s ↓    & CHAIR\_i ↓    & Pre.          & F1            \\ \hline
Baseline                   & 50.2          & 15.4          & 72.1          & 73.5          \\
MaskCD                     & \textbf{40.6} & \textbf{10.8} & \textbf{79.1} & \textbf{78.2} \\
MaskCD\_r                  & 44.0          & 14.3          & 77.3          & 74.0          \\ \hline
\end{tabular}%
}
\caption{\textbf{Results of MaskCD and MaskCD\_r on CHAIR evaluation.} "Pre." is the abbreviation for "Precision". It is evident that MaskCD\_r indeed helps mitigate hallucination, but cannot compete with MaskCD.}
\label{tab:ablation-chair}
\end{table}

\begin{table}[!t]
\centering
\resizebox{\columnwidth}{!}{%
\begin{tabular}{clcllc}
\hline
                              & \multicolumn{1}{c}{Method} & Accuracy       & Precision      & Recall         & F1             \\ \hline
\multirow{3}{*}{\textit{All}} & Baseline                   & 80.87          & 87.62          & 72.07          & 79.05          \\
                              & MaskCD                     & \textbf{84.88} & 83.11          & \textbf{88.20} & \textbf{85.48} \\
                              & MaskCD\_r                  & 83.61          & \textbf{90.67} & 75.53          & 82.38          \\ \hline
\end{tabular}%
}
\caption{\textbf{Results of MaskCD and MaskCD\_r on POPE benchmark.} MaskCD\_r performs best on Precision metrics, but fails in Recall.}
\label{tab:ablation pope}
\end{table}

\begin{table}[!ht]
\centering
\resizebox{\columnwidth}{!}{%
\begin{tabular}{lcccc}
\hline
\multicolumn{1}{c}{\multirow{2}{*}{Method}} & \multicolumn{2}{c}{Object-level} & \multicolumn{2}{c}{Attribute-level} \\ \cline{2-5} 
\multicolumn{1}{c}{} & existence       & count           & position        & color           \\ \hline
Baseline             & 180.00          & 101.67          & 100.00          & \textbf{153.33} \\
MaskCD               & \textbf{195.00} & \textbf{168.33} & \textbf{133.33} & 150.00          \\
MaskCD\_r            & 190.00          & 133.33          & 123.33          & 150.00          \\ \hline
\end{tabular}%
}
\caption{\textbf{Results of MaskCD and MaskCD\_r on the hallucination-related subsets of MME.} MaskCD\_r indeed helps slightly in mitigating hallucinations.}
\label{tab:ablation-mme}
\end{table}

\paragraph{Mask proportion and CD tensity} As mentioned in section \ref{subsec:mask formulation} and \ref{subsec: maskcd}, there are two important hyper-parameters in MaskCD: $\tau$, as the threshold for determining the image head, controls the number of masked heads; and $\alpha$, which controls the intensity of contrastive decoding operation. 

We conduct ablation experiments of $\tau$ and $\alpha$ on LLaVA-1.5-7b with CHAIR evaluation. Table \ref{tab:ablation-tau-chair} shows the results of MaskCD with different thresholds $\tau$. It indicates that the best value of $\tau$ is 0.9, which means around 23\% of the heads in LLaVA-1.5-7b's LLM backbone are recognized as image heads and have been masked (according to Table \ref{tab:head num}). Whether too small or too big, the value of $\tau$ is, the performances tend to decline, and even fail the baseline when $\tau$ is 0.5. This shows that the heads to be masked should be delicately selected, and MaskCD achieves this successfully.

Meanwhile, Table \ref{tab:ablation-alpha-chair} shows the results of MaskCD on CHAIR with different $\alpha$. $\alpha$ is a common hyperparameter in contrastive decoding approaches, whose value controls the intensity of the contrast operation. It can be seen that even when the value of $\alpha$ is quite large, MaskCD can still operate stably and effectively alleviate the hallucination phenomenon. It demonstrates the stability, reliability and practicability of MaskCD as a CD method.


\begin{table}[!t]
\centering
\resizebox{\columnwidth}{!}{%
\begin{tabular}{ccccc}
\hline
Method                  & $\tau$ & CHAIR\_s      & CHAIR\_i      & F1            \\ \hline
Baseline                & /                   & 50.2          & 15.4          & 73.5          \\
\multirow{6}{*}{MaskCD} & 0.95                & 46.8          & 12.9          & 77.0          \\
                        & 0.9                 & \textbf{40.6} & \textbf{10.8} & \textbf{78.2} \\
                        & 0.8                 & 40.8          & 14.7          & 74.8          \\
                        & 0.7                 & 48.4          & 13.0          & 76.4          \\
                        & 0.6                 & 49.6          & 14.0          & 75.9          \\
                        & 0.5                 & 54.8          & 14.4          & 75.2          \\ \hline
\end{tabular}%
}
\caption{\textbf{Results of MaskCD with different threshold $\tau$.} It can be seen that the value of $\tau$ that is either too small or too large is not conducive to dealing with hallucination problems.}
\label{tab:ablation-tau-chair}
\end{table}

\begin{table}[!t]
\centering
\resizebox{\columnwidth}{!}{%
\begin{tabular}{ccccc}
\hline
Method                  & $\alpha$ & CHAIR\_s      & CHAIR\_i      & F1            \\ \hline
Baseline                & /        & 50.2          & 15.4          & 73.5          \\
\multirow{7}{*}{MaskCD} & 0.5      & 43.6          & 11.5          & \textbf{78.4} \\
                        & 1.0      & \textbf{40.6} & \textbf{10.8} & 78.2          \\
                        & 2.0      & 45.2          & 12.2          & 77.3          \\
                        & 3.0      & 44.6          & 11.6          & 77.3          \\
                        & 4.0      & 42.2          & 11.5          & 77.6          \\
                        & 5.0      & 41.6          & 11.7          & 77.9          \\
                        & 6.0      & 41.2          & 11.8          & 77.9          \\ \hline
\end{tabular}%
}
\caption{\textbf{Results of MaskCD with different $\alpha$. } It can be seen that even when $\alpha$ takes a large value, MaskCD can still operate stably, effectively alleviating the hallucination phenomenon.}
\label{tab:ablation-alpha-chair}
\end{table}

\section{Conclusion}
In this paper, we first introduce the image heads: the heads in LVLM's LLM backbone that tend to pay comparably high attention proportion on image tokens. Then we propose the image head Masked Contrastive Decoding (MaskCD) method, a novel contrastive decoding approach featuring in masking image heads to construct contrastive samples. MaskCD constructs the contrastive samples of CD methods from the perspective of attention score, combining the effectiveness and stability of these two methods. Extensive experimental results on CHAIR, POPE and MME demonstrate the effectiveness and stability of MaskCD in mitigating the phenomenon of hallucinations. We hope this work can provide a new perspective for exploring future efforts to alleviate hallucinations in LVLMs.

\section*{Limitations}

Although MaskCD achieves significant performance in hallucination mitigation, it still has several limitations. First, MaskCD requires the use of images for inference in advance to obtain the masks of image heads, which occupy computing resources. Secondly, although the process of obtaining the mask is simple, the obtained mask is only applicable to the same family of LLM backbones. For new LLM bases, the corresponding masks need to be re-obtained. This limitation encourages us to explore how to dynamically construct masks during the model's operation, so as to get rid of these restrictions.

\section*{Ethical Considerations}
The main research objects of this work are alleviating hallucination phenomenon, which help avoid disloyal contents generated by LVLMs.  Moreover, we conduct experiments on the public datasets, which do not contain any offensive content or information with negative social impact.
Our research contents are completely in line with the ethical review.

\section*{Acknowledgments}
This work was partly supported by the National Key Research and Development Program of China (No. 2024YFB2808903) and the  research grant No. CT20240905126002 of the Doubao Large Model Fund.

\bibliography{custom}

\begin{thebibliography}{37}
\providecommand{\natexlab}[1]{#1}

\bibitem[{Alayrac et~al.(2022)Alayrac, Donahue, Luc, Miech, Barr, Hasson, Lenc, Mensch, Millican, Reynolds, Ring, Rutherford, Cabi, Han, Gong, Samangooei, Monteiro, Menick, Borgeaud, Brock, Nematzadeh, Sharifzadeh, Binkowski, Barreira, Vinyals, Zisserman, and Simonyan}]{DBLP:conf/nips/AlayracDLMBHLMM22}
Jean{-}Baptiste Alayrac, Jeff Donahue, Pauline Luc, Antoine Miech, Iain Barr, Yana Hasson, Karel Lenc, Arthur Mensch, Katherine Millican, Malcolm Reynolds, Roman Ring, Eliza Rutherford, Serkan Cabi, Tengda Han, Zhitao Gong, Sina Samangooei, Marianne Monteiro, Jacob~L. Menick, Sebastian Borgeaud, and 8 others. 2022.
\newblock Flamingo: a visual language model for few-shot learning.
\newblock In \emph{NeurIPS}.

\bibitem[{Bai et~al.(2023{\natexlab{a}})Bai, Bai, Chu, Cui, Dang, Deng, Fan, Ge, Han, Huang, Hui, Ji, Li, Lin, Lin, Liu, Liu, Lu, Lu, Ma, Men, Ren, Ren, Tan, Tan, Tu, Wang, Wang, Wang, Wu, Xu, Xu, Yang, Yang, Yang, Yang, Yao, Yu, Yuan, Yuan, Zhang, Zhang, Zhang, Zhang, Zhou, Zhou, Zhou, and Zhu}]{DBLP:journals/corr/abs-2309-16609}
Jinze Bai, Shuai Bai, Yunfei Chu, Zeyu Cui, Kai Dang, Xiaodong Deng, Yang Fan, Wenbin Ge, Yu~Han, Fei Huang, Binyuan Hui, Luo Ji, Mei Li, Junyang Lin, Runji Lin, Dayiheng Liu, Gao Liu, Chengqiang Lu, Keming Lu, and 29 others. 2023{\natexlab{a}}.
\newblock Qwen technical report.
\newblock \emph{CoRR}, abs/2309.16609.

\bibitem[{Bai et~al.(2023{\natexlab{b}})Bai, Bai, Yang, Wang, Tan, Wang, Lin, Zhou, and Zhou}]{DBLP:journals/corr/abs-2308-12966}
Jinze Bai, Shuai Bai, Shusheng Yang, Shijie Wang, Sinan Tan, Peng Wang, Junyang Lin, Chang Zhou, and Jingren Zhou. 2023{\natexlab{b}}.
\newblock Qwen-vl: {A} frontier large vision-language model with versatile abilities.
\newblock \emph{CoRR}, abs/2308.12966.

\bibitem[{Brown et~al.(2020)Brown, Mann, Ryder, Subbiah, Kaplan, Dhariwal, Neelakantan, Shyam, Sastry, Askell, Agarwal, Herbert{-}Voss, Krueger, Henighan, Child, Ramesh, Ziegler, Wu, Winter, Hesse, Chen, Sigler, Litwin, Gray, Chess, Clark, Berner, McCandlish, Radford, Sutskever, and Amodei}]{DBLP:conf/nips/BrownMRSKDNSSAA20}
Tom~B. Brown, Benjamin Mann, Nick Ryder, Melanie Subbiah, Jared Kaplan, Prafulla Dhariwal, Arvind Neelakantan, Pranav Shyam, Girish Sastry, Amanda Askell, Sandhini Agarwal, Ariel Herbert{-}Voss, Gretchen Krueger, Tom Henighan, Rewon Child, Aditya Ramesh, Daniel~M. Ziegler, Jeffrey Wu, Clemens Winter, and 12 others. 2020.
\newblock \href {https://proceedings.neurips.cc/paper/2020/hash/1457c0d6bfcb4967418bfb8ac142f64a-Abstract.html} {Language models are few-shot learners}.
\newblock In \emph{Advances in Neural Information Processing Systems 33: Annual Conference on Neural Information Processing Systems 2020, NeurIPS 2020, December 6-12, 2020, virtual}.

\bibitem[{Chen et~al.(2024)Chen, Sikka, Cogswell, Ji, and Divakaran}]{DBLP:conf/cvpr/ChenSCJD24}
Yangyi Chen, Karan Sikka, Michael Cogswell, Heng Ji, and Ajay Divakaran. 2024.
\newblock {DRESS} : Instructing large vision-language models to align and interact with humans via natural language feedback.
\newblock In \emph{{CVPR}}, pages 14239--14250. {IEEE}.

\bibitem[{Chiang et~al.(2023)Chiang, Li, Lin, Sheng, Wu, Zhang, Zheng, Siyuan, Zhuang, Zhuang, E, Gonzalez, Stoica, and Xing.}]{vicuna_blog}
Wei-Lin Chiang, Zhuohan Li, Zi~Lin, Ying Sheng, Zhanghao Wu, Hao Zhang, Lianmin Zheng, Siyuan, Zhuang, Yonghao Zhuang, Joseph E, Gonzalez, Ion Stoica, and Eric~P. Xing. 2023.
\newblock \href {https://lmsys.org/blog/2023-03-30-vicuna/} {Vicuna: An open-source chatbot impressing gpt-4 with 90\%* chatgpt quality}.
\newblock Unpublished blog post.

\bibitem[{Dai et~al.(2023)Dai, Li, Li, Tiong, Zhao, Wang, Li, Fung, and Hoi}]{DBLP:conf/nips/Dai0LTZW0FH23}
Wenliang Dai, Junnan Li, Dongxu Li, Anthony Meng~Huat Tiong, Junqi Zhao, Weisheng Wang, Boyang Li, Pascale Fung, and Steven C.~H. Hoi. 2023.
\newblock Instructblip: Towards general-purpose vision-language models with instruction tuning.
\newblock In \emph{NeurIPS}.

\bibitem[{Favero et~al.(2024)Favero, Zancato, Trager, Choudhary, Perera, Achille, Swaminathan, and Soatto}]{DBLP:conf/cvpr/FaveroZTCPASS24}
Alessandro Favero, Luca Zancato, Matthew Trager, Siddharth Choudhary, Pramuditha Perera, Alessandro Achille, Ashwin Swaminathan, and Stefano Soatto. 2024.
\newblock Multi-modal hallucination control by visual information grounding.
\newblock In \emph{{CVPR}}, pages 14303--14312. {IEEE}.

\bibitem[{Fu et~al.(2023)Fu, Chen, Shen, Qin, Zhang, Lin, Qiu, Lin, Yang, Zheng, Li, Sun, and Ji}]{DBLP:journals/corr/abs-2306-13394}
Chaoyou Fu, Peixian Chen, Yunhang Shen, Yulei Qin, Mengdan Zhang, Xu~Lin, Zhenyu Qiu, Wei Lin, Jinrui Yang, Xiawu Zheng, Ke~Li, Xing Sun, and Rongrong Ji. 2023.
\newblock {MME:} {A} comprehensive evaluation benchmark for multimodal large language models.
\newblock \emph{CoRR}, abs/2306.13394.

\bibitem[{Hu et~al.(2023)Hu, Zhang, Zhao, and Sun}]{DBLP:journals/corr/abs-2309-02301}
Hongyu Hu, Jiyuan Zhang, Minyi Zhao, and Zhenbang Sun. 2023.
\newblock {CIEM:} contrastive instruction evaluation method for better instruction tuning.
\newblock \emph{CoRR}, abs/2309.02301.

\bibitem[{Huang et~al.(2024)Huang, Dong, Zhang, Wang, He, Wang, Lin, Zhang, and Yu}]{DBLP:conf/cvpr/HuangDZ0H0L0Y24}
Qidong Huang, Xiaoyi Dong, Pan Zhang, Bin Wang, Conghui He, Jiaqi Wang, Dahua Lin, Weiming Zhang, and Nenghai Yu. 2024.
\newblock {OPERA:} alleviating hallucination in multi-modal large language models via over-trust penalty and retrospection-allocation.
\newblock In \emph{{CVPR}}, pages 13418--13427. {IEEE}.

\bibitem[{Kamath et~al.(2023)Kamath, Hessel, and Chang}]{DBLP:conf/emnlp/KamathHC23a}
Amita Kamath, Jack Hessel, and Kai{-}Wei Chang. 2023.
\newblock What's "up" with vision-language models? investigating their struggle with spatial reasoning.
\newblock In \emph{{EMNLP}}, pages 9161--9175. Association for Computational Linguistics.

\bibitem[{Lee et~al.(2024)Lee, Park, Jo, and Seo}]{DBLP:conf/naacl/LeePJS24}
Seongyun Lee, Sue~Hyun Park, Yongrae Jo, and Minjoon Seo. 2024.
\newblock Volcano: Mitigating multimodal hallucination through self-feedback guided revision.
\newblock In \emph{{NAACL-HLT}}, pages 391--404. Association for Computational Linguistics.

\bibitem[{Leng et~al.(2024)Leng, Zhang, Chen, Li, Lu, Miao, and Bing}]{DBLP:conf/cvpr/LengZCLLMB24}
Sicong Leng, Hang Zhang, Guanzheng Chen, Xin Li, Shijian Lu, Chunyan Miao, and Lidong Bing. 2024.
\newblock Mitigating object hallucinations in large vision-language models through visual contrastive decoding.
\newblock In \emph{{CVPR}}, pages 13872--13882. {IEEE}.

\bibitem[{Li et~al.(2023{\natexlab{a}})Li, Li, Savarese, and Hoi}]{DBLP:conf/icml/0008LSH23}
Junnan Li, Dongxu Li, Silvio Savarese, and Steven C.~H. Hoi. 2023{\natexlab{a}}.
\newblock {BLIP-2:} bootstrapping language-image pre-training with frozen image encoders and large language models.
\newblock In \emph{{ICML}}, volume 202 of \emph{Proceedings of Machine Learning Research}, pages 19730--19742. {PMLR}.

\bibitem[{Li et~al.(2023{\natexlab{b}})Li, Du, Zhou, Wang, Zhao, and Wen}]{DBLP:conf/emnlp/LiDZWZW23}
Yifan Li, Yifan Du, Kun Zhou, Jinpeng Wang, Wayne~Xin Zhao, and Ji{-}Rong Wen. 2023{\natexlab{b}}.
\newblock Evaluating object hallucination in large vision-language models.
\newblock In \emph{{EMNLP}}, pages 292--305. Association for Computational Linguistics.

\bibitem[{Lin et~al.(2014)Lin, Maire, Belongie, Hays, Perona, Ramanan, Doll{\'{a}}r, and Zitnick}]{DBLP:conf/eccv/LinMBHPRDZ14}
Tsung{-}Yi Lin, Michael Maire, Serge~J. Belongie, James Hays, Pietro Perona, Deva Ramanan, Piotr Doll{\'{a}}r, and C.~Lawrence Zitnick. 2014.
\newblock Microsoft {COCO:} common objects in context.
\newblock In \emph{{ECCV} {(5)}}, volume 8693 of \emph{Lecture Notes in Computer Science}, pages 740--755. Springer.

\bibitem[{Liu et~al.(2023{\natexlab{a}})Liu, Lin, Li, Wang, Yacoob, and Wang}]{DBLP:journals/corr/abs-2306-14565}
Fuxiao Liu, Kevin Lin, Linjie Li, Jianfeng Wang, Yaser Yacoob, and Lijuan Wang. 2023{\natexlab{a}}.
\newblock Aligning large multi-modal model with robust instruction tuning.
\newblock \emph{CoRR}, abs/2306.14565.

\bibitem[{Liu et~al.(2024{\natexlab{a}})Liu, Lin, Li, Wang, Yacoob, and Wang}]{DBLP:conf/iclr/LiuLLWYW24}
Fuxiao Liu, Kevin Lin, Linjie Li, Jianfeng Wang, Yaser Yacoob, and Lijuan Wang. 2024{\natexlab{a}}.
\newblock Mitigating hallucination in large multi-modal models via robust instruction tuning.
\newblock In \emph{{ICLR}}. OpenReview.net.

\bibitem[{Liu et~al.(2023{\natexlab{b}})Liu, Li, Li, and Lee}]{DBLP:journals/corr/abs-2310-03744}
Haotian Liu, Chunyuan Li, Yuheng Li, and Yong~Jae Lee. 2023{\natexlab{b}}.
\newblock Improved baselines with visual instruction tuning.
\newblock \emph{CoRR}, abs/2310.03744.

\bibitem[{Liu et~al.(2024{\natexlab{b}})Liu, Zheng, and Chen}]{DBLP:conf/eccv/LiuZC24}
Shi Liu, Kecheng Zheng, and Wei Chen. 2024{\natexlab{b}}.
\newblock Paying more attention to image: {A} training-free method for alleviating hallucination in lvlms.
\newblock In \emph{{ECCV} {(83)}}, volume 15141 of \emph{Lecture Notes in Computer Science}, pages 125--140. Springer.

\bibitem[{Nie et~al.(2024)Nie, Zhang, An, Tan, Kot, and Lu}]{DBLP:journals/corr/abs-2406-09121}
Jiahao Nie, Gongjie Zhang, Wenbin An, Yap{-}Peng Tan, Alex~C. Kot, and Shijian Lu. 2024.
\newblock Mmrel: {A} relation understanding dataset and benchmark in the {MLLM} era.
\newblock \emph{CoRR}, abs/2406.09121.

\bibitem[{OpenAI(2023)}]{DBLP:journals/corr/abs-2303-08774}
OpenAI. 2023.
\newblock \href {https://doi.org/10.48550/ARXIV.2303.08774} {{GPT-4} technical report}.
\newblock \emph{CoRR}, abs/2303.08774.

\bibitem[{Radford et~al.(2021)Radford, Kim, Hallacy, Ramesh, Goh, Agarwal, Sastry, Askell, Mishkin, Clark, Krueger, and Sutskever}]{DBLP:conf/icml/RadfordKHRGASAM21}
Alec Radford, Jong~Wook Kim, Chris Hallacy, Aditya Ramesh, Gabriel Goh, Sandhini Agarwal, Girish Sastry, Amanda Askell, Pamela Mishkin, Jack Clark, Gretchen Krueger, and Ilya Sutskever. 2021.
\newblock Learning transferable visual models from natural language supervision.
\newblock In \emph{{ICML}}, volume 139 of \emph{Proceedings of Machine Learning Research}, pages 8748--8763. {PMLR}.

\bibitem[{Rohrbach et~al.(2018)Rohrbach, Hendricks, Burns, Darrell, and Saenko}]{DBLP:conf/emnlp/RohrbachHBDS18}
Anna Rohrbach, Lisa~Anne Hendricks, Kaylee Burns, Trevor Darrell, and Kate Saenko. 2018.
\newblock Object hallucination in image captioning.
\newblock In \emph{{EMNLP}}, pages 4035--4045. Association for Computational Linguistics.

\bibitem[{Sun et~al.(2024)Sun, Shen, Cao, Liu, Li, Shen, Gan, Gui, Wang, Yang, Keutzer, and Darrell}]{DBLP:conf/acl/SunSCLLSGGWYKD24}
Zhiqing Sun, Sheng Shen, Shengcao Cao, Haotian Liu, Chunyuan Li, Yikang Shen, Chuang Gan, Liangyan Gui, Yu{-}Xiong Wang, Yiming Yang, Kurt Keutzer, and Trevor Darrell. 2024.
\newblock Aligning large multimodal models with factually augmented {RLHF}.
\newblock In \emph{{ACL} (Findings)}, pages 13088--13110. Association for Computational Linguistics.

\bibitem[{Touvron et~al.(2023)Touvron, Lavril, Izacard, Martinet, Lachaux, Lacroix, Rozi{\`{e}}re, Goyal, Hambro, Azhar, Rodriguez, Joulin, Grave, and Lample}]{DBLP:journals/corr/abs-2302-13971}
Hugo Touvron, Thibaut Lavril, Gautier Izacard, Xavier Martinet, Marie{-}Anne Lachaux, Timoth{\'{e}}e Lacroix, Baptiste Rozi{\`{e}}re, Naman Goyal, Eric Hambro, Faisal Azhar, Aur{\'{e}}lien Rodriguez, Armand Joulin, Edouard Grave, and Guillaume Lample. 2023.
\newblock Llama: Open and efficient foundation language models.
\newblock \emph{CoRR}, abs/2302.13971.

\bibitem[{Tu et~al.(2025)Tu, Ye, Zhou, Bai, Yu, Chen, and Ouyang}]{DBLP:journals/corr/abs-2503-08342}
Chongjun Tu, Peng Ye, Dongzhan Zhou, Lei Bai, Gang Yu, Tao Chen, and Wanli Ouyang. 2025.
\newblock Attention reallocation: Towards zero-cost and controllable hallucination mitigation of mllms.
\newblock \emph{CoRR}, abs/2503.08342.

\bibitem[{Wang et~al.(2023)Wang, Wang, Xu, Zhang, Gu, Jia, Wang, Xu, Yan, Zhang et~al.}]{wang2023amber}
Junyang Wang, Yuhang Wang, Guohai Xu, Jing Zhang, Yukai Gu, Haitao Jia, Jiaqi Wang, Haiyang Xu, Ming Yan, Ji~Zhang, and 1 others. 2023.
\newblock Amber: An llm-free multi-dimensional benchmark for mllms hallucination evaluation.
\newblock \emph{arXiv preprint arXiv:2311.07397}.

\bibitem[{Wang et~al.(2024)Wang, Pan, Ding, and Biemann}]{DBLP:conf/acl/WangPDB24}
Xintong Wang, Jingheng Pan, Liang Ding, and Chris Biemann. 2024.
\newblock Mitigating hallucinations in large vision-language models with instruction contrastive decoding.
\newblock In \emph{{ACL} (Findings)}, pages 15840--15853. Association for Computational Linguistics.

\bibitem[{Woo et~al.(2024)Woo, Jang, Kim, Choi, and Kim}]{DBLP:journals/corr/abs-2405-17821}
Sangmin Woo, Jaehyuk Jang, Donguk Kim, Yubin Choi, and Changick Kim. 2024.
\newblock {RITUAL:} random image transformations as a universal anti-hallucination lever in lvlms.
\newblock \emph{CoRR}, abs/2405.17821.

\bibitem[{Xiao et~al.(2024)Xiao, Tian, Chen, Han, and Lewis}]{DBLP:conf/iclr/XiaoTCHL24}
Guangxuan Xiao, Yuandong Tian, Beidi Chen, Song Han, and Mike Lewis. 2024.
\newblock Efficient streaming language models with attention sinks.
\newblock In \emph{{ICLR}}. OpenReview.net.

\bibitem[{Ye et~al.(2023)Ye, Xu, Ye, Yan, Hu, Liu, Qian, Zhang, Huang, and Zhou}]{DBLP:journals/corr/abs-2311-04257}
Qinghao Ye, Haiyang Xu, Jiabo Ye, Ming Yan, Anwen Hu, Haowei Liu, Qi~Qian, Ji~Zhang, Fei Huang, and Jingren Zhou. 2023.
\newblock mplug-owl2: Revolutionizing multi-modal large language model with modality collaboration.
\newblock \emph{CoRR}, abs/2311.04257.

\bibitem[{Yin et~al.(2023)Yin, Fu, Zhao, Xu, Wang, Sui, Shen, Li, Sun, and Chen}]{DBLP:journals/corr/abs-2310-16045}
Shukang Yin, Chaoyou Fu, Sirui Zhao, Tong Xu, Hao Wang, Dianbo Sui, Yunhang Shen, Ke~Li, Xing Sun, and Enhong Chen. 2023.
\newblock Woodpecker: Hallucination correction for multimodal large language models.
\newblock \emph{CoRR}, abs/2310.16045.

\bibitem[{Zhang et~al.(2024)Zhang, Yu, Wen, Wang, Zhang, Wang, Jin, and Tan}]{DBLP:journals/corr/abs-2403-05262}
Yifan Zhang, Weichen Yu, Qingsong Wen, Xue Wang, Zhang Zhang, Liang Wang, Rong Jin, and Tieniu Tan. 2024.
\newblock Debiasing multimodal large language models.
\newblock \emph{CoRR}, abs/2403.05262.

\bibitem[{Zhu et~al.(2023)Zhu, Chen, Shen, Li, and Elhoseiny}]{DBLP:journals/corr/abs-2304-10592}
Deyao Zhu, Jun Chen, Xiaoqian Shen, Xiang Li, and Mohamed Elhoseiny. 2023.
\newblock Minigpt-4: Enhancing vision-language understanding with advanced large language models.
\newblock \emph{CoRR}, abs/2304.10592.

\bibitem[{Zhu et~al.(2024)Zhu, Ji, Chen, Xu, Ye, and Liu}]{DBLP:journals/corr/abs-2402-18476}
Lanyun Zhu, Deyi Ji, Tianrun Chen, Peng Xu, Jieping Ye, and Jun Liu. 2024.
\newblock {IBD:} alleviating hallucinations in large vision-language models via image-biased decoding.
\newblock \emph{CoRR}, abs/2402.18476.

\end{thebibliography}

\appendix

\section{Details about image head selection}
\label{sec:appendix}

\subsection{The Formalized Description of MaskCD}
In short, MaskCD is a type of hallucination mitigation method that constructs negative samples by masking the heads in the LLM backbone that pay high attention to image information. It follows the steps below:

Generate captions using test cases:

\begin{equation}
  \label{eq:app1}
  \begin{split}
    p_t = \text{softmax}( & \text{logits}_\theta (y | y_{<t}, X_{system}, \\
                          & X_{image}, X_{instruction}) )
  \end{split}
\end{equation}

In which $X_{system}$, $X_{image}$ and $X_{instruction}$ represent the tokens of system prompt, image and instruction prompt, respectively.

For each output token generation, we compute the sum of attention scores received by the image tokens from each attention head. This forms an image-token attention score matrix for that specific token generation. After all output tokens for all images have been generated, we obtain a total of \textit{num\_of\_total\_tokens} attention score matrices, each with the shape [num\_of\_layers, num\_of\_heads].

\begin{equation}
  \label{eq:app2}
  \begin{split}
    A & \in \mathbb{R}^{T\times L\times H}, \\
    A_{t,i,j} & = \sum_{X_k\in X_{image}} \text{AttentionScore}_{i,j,k}[k]
  \end{split}
\end{equation}

Where $A_{t,i,j}$ denotes the sum of attention scores received by the image tokens from the 
j-th head in the i-th layer at time step t. T represents the total number of generated tokens, L is the number of attention layers in the LLM, and H is the number of attention heads per layer.

Then, for all the collected data, we determine whether each value exceeds a threshold $\tau$; each time it does, we increment a count by 1. In this way, we obtain statistical data on the attention heads whose attention to image tokens exceeds the threshold:

\begin{equation}
  \label{eq:app3}
  \begin{split}
    C & \in \mathbb{R}^{L\times H}, \\
    C_{i,j} & = \sum^{T}_{t=1} \mathbf{1}_{A_{t,i,j}>\tau}
  \end{split}
\end{equation}

The resulting matrix still has the shape[num\_of\_layers,num\_of\_heads], where each element represents the number of times the attention score of the corresponding head exceeded the threshold $\tau$ during all T generation steps.

At this point, all positions with non-zero values are identified as image heads. These positions are set to 0, while all other positions are set to 1, thereby forming the masking matrix:

\begin{equation}
  \label{eq:app4}
  \begin{split}
    \text{ImageHeadMask} & \in \mathbb{R}^{L\times H}, \\
    \text{ImageHeadMask}_{i,j} & = \begin{cases}
        1, & C_{i,j}=0 \\
        0, & C_{i,j}>0
    \end{cases}
  \end{split}
\end{equation}
  
During the inference phase, when generating logits for bad samples, the hidden state at each layer is multiplied by the corresponding layer's IHM (Image Head Mask) before being output, thereby masking the image heads. The original samples remain unaffected. Finally, the logits from both are compared, and only the final logits are used for output generation.

\subsection{Influencing factors to image head selection}

\paragraph{Impact of image types.} We sampled 500 images from each of the COCO2014-val, MMRel, ArtBench, and ChartQA datasets for the captioning task, representing real-world images, AI-generated images, artistic paintings, and chart-based visualizations, respectively. The first three primarily differ in visual style, while the last focuses on the graphical representation of textual and numerical information. In addition, we sampled 125 images from each of the four datasets to construct a mixed image set of 500 samples.

The evaluation results are presented in the following two tables. Table \ref{tab:h1} shows the number of image heads selected under different threshold values for each image type. As can be observed, the number of selected image heads is generally similar across different types of images, except for AI-generated images, which tend to yield slightly fewer image heads at higher thresholds.

\begin{table*}[]
\centering
\resizebox{\textwidth}{!}{
\begin{tabular}{lllllllll}
\hline
image type        & $\tau$=0.95 & $\tau$=0.9 & $\tau$=0.8 & $\tau$=0.7 & $\tau$=0.6 & $\tau$=0.5 & total token number & average length \\ \hline
COCO-RealWorld    & 192    & 238   & 315   & 364   & 424   & 506   & 65501184           & 127.93         \\
ArtBench-painting & 175    & 222   & 294   & 339   & 390   & 466   & 66680832           & 130.24         \\
MMrel-AIGC        & 161    & 219   & 285   & 364   & 424   & 506   & 59910505           & 117.01         \\
ChartQA-chart     & 187    & 229   & 307   & 355   & 419   & 493   & 66335744           & 129.56         \\
Mixture           & 194    & 238   & 304   & 356   & 416   & 494   & 32612352           & 127.39         \\ \hline
\end{tabular}%
}
\caption{}
\label{tab:h1}
\end{table*}

Table \ref{tab:h2} shows the overlap between the image heads selected from the real-image set and those from the other four image sets. For each entry, we report three values: the size of the intersection, the size of the union, and the overlap ratio (i.e., intersection size divided by union size). As shown, the attention heads selected across different image types exhibit a high degree of overlap with those from real images. This indicates that our current method for identifying image heads remains relatively stable across various image domains.

\begin{table*}[]
\centering
\resizebox{\textwidth}{!}{
\begin{tabular}{lllllll}
\hline
image type & $\tau$=0.95       & $\tau$=0.9        & $\tau$=0.8        & $\tau$=0.7        & $\tau$=0.6        & $\tau$=0.5        \\ \hline
ArtBench-painting & 170/197/0.86 & 219/241/0.91 & 292/317/0.92 & 333/370/0.90 & 380/434/0.88 & 459/513/0.89 \\
ChartQA-chart     & 166/213/0.78 & 203/264/0.77 & 280/342/0.82 & 325/394/0.82 & 375/468/0.80 & 445/554/0.80 \\
MMrel-AIGC & 150/203/0.74 & 202/255/0.79 & 274/326/0.84 & 364/364/1.00 & 424/424/1.00 & 506/506/1.00 \\
Mixture    & 175/211/0.83 & 216/260/0.83 & 289/330/0.88 & 335/385/0.87 & 383/457/0.84 & 459/541/0.85 \\ \hline
\end{tabular}%
}
\caption{}
\label{tab:h2}
\end{table*}

\paragraph{Impact of image quantity.} We selected 1,000 real-world images from the COCO2014-val dataset and conducted captioning task. During the process, we recorded the selected image heads after 100, 300, 500, and 1,000 captions, in order to avoid evaluation differences caused by varying image content.

The test results are shown in the following two tables. Table \ref{tab:h3} presents the number of selected image heads under different thresholds and image quantities. It shows that the number of image heads stabilizes after 300–500 images.

\begin{table*}[]
\centering
\begin{tabular}{lllllllll}
\hline
card & 0.95 & 0.9 & 0.8 & 0.7 & 0.6 & 0.5 & total token number & average len \\ \hline
100  & 178  & 230 & 294 & 342 & 391 & 462 & 12836864           & 125.36      \\
300  & 198  & 250 & 314 & 362 & 411 & 482 & 39139328           & 127.41      \\
500  & 192  & 238 & 315 & 364 & 424 & 506 & 65501184           & 127.93      \\
1000 & 202  & 248 & 325 & 374 & 434 & 516 & 129914880          & 126.87      \\ \hline
\end{tabular}%
\caption{}
\label{tab:h3}
\end{table*}

Table \ref{tab:h4} presents the overlap between the image heads selected using 100, 300, and 500 images, compared to the 1,000-image result. This demonstrates that as the number of test images increases, the distribution of selected image heads remains highly stable, with only minor increases in total count.

\begin{table*}[]
\centering
\begin{tabular}{lllllll}
\hline
card & 0.95         & 0.9          & 0.8          & 0.7          & 0.6          & 0.5          \\ \hline
100  & 178/202/0.88 & 230/248/0.93 & 294/325/0.90 & 342/374/0.91 & 391/434/0.90 & 462/516/0.90 \\
300  & 187/202/0.93 & 236/248/0.95 & 309/325/0.95 & 353/374/0.94 & 406/434/0.94 & 475/516/0.92 \\
500  & 189/202/0.94 & 237/248/0.96 & 309/325/0.95 & 356/374/0.95 & 408/434/0.94 & 487/516/0.94 \\ \hline
\end{tabular}%
\caption{}
\label{tab:h4}
\end{table*}

\paragraph{Impact of task type.} We decided to compare captioning and discriminative VQA tasks. Since the amount of data collected each time depends on the total number of generated tokens, and the generation lengths differ significantly between captioning and discriminative VQA tasks, we adjusted the number of samples to balance the total generated tokens. Specifically, we designed two setups: 100 real images for captioning with a generation length limited to 20 tokens versus 1000 VQA tasks, and 100 real images for captioning without length restriction versus 6000 VQA tasks.

The results of the two tests are shown in table \ref{tab:h5}. We found that for the VQA task, since the model can only generate “yes” or “no” answers, the number of indicated image heads is very limited. Inspired by this phenomenon, we conducted a simple investigation on the number of image heads indicated by different semantic token types in caption generation. We found that attribute tokens such as colors indicate more image heads, nouns indicate fewer, and other tokens like “yes,” “no,” or “the” indicate significantly fewer heads. This phenomenon may be explained by the fact that during the generation of attribute and noun tokens, the model needs to rely more heavily on visual information, which increases the number of image heads collected.

\begin{table*}[]
\centering
\begin{tabular}{llllllllll}
\hline
card & task                   & 0.95 & 0.9 & 0.8 & 0.7 & 0.6 & 0.5 & token      & len    \\ \hline
100  & caption (maxlength 20) & 133  & 184 & 245 & 297 & 336 & 406 & 2,048,000  & 20.00  \\
1000 & vqa                    & 1    & 8   & 35  & 44  & 66  & 81  & 2,148,352  & 2.10   \\
100  & caption (no limit)     & 178  & 230 & 294 & 342 & 391 & 462 & 12,836,864 & 125.36 \\
6000 & vqa                    & 4    & 16  & 66  & 108 & 149 & 189 & 12,657,664 & 2.06   \\ \hline
\end{tabular}%
\caption{}
\label{tab:h5}
\end{table*}

\end{document}